\documentclass[runningheads]{llncs}
\usepackage{amsmath,amssymb}
\DeclareMathOperator{\E}{\mathbb{E}}
\usepackage{graphicx}
\usepackage{multirow}
\usepackage{makecell}
\usepackage{array}

\usepackage{diagbox}
\newcommand{\thickcline}[1]{%
    \@thickcline #1\@nil%
}

\def\@thickcline#1-#2\@nil{%
  \omit
  \@multicnt#1%
  \advance\@multispan\m@ne
  \ifnum\@multicnt=\@ne\@firstofone{&\omit}\fi
  \@multicnt#2%
  \advance\@multicnt-#1%
  \advance\@multispan\@ne
  \leaders\hrule\@height1pt\hfill
  \cr
  \noalign{\vskip-1pt}%
}

\usepackage[symbol]{footmisc}

\newcommand*\samethanks[1][\value{footnote}]{\footnotemark[#1]}

\newcommand{\norm}[1]{\left\lVert#1\right\rVert}
\begin{document}
\title{Co-Generation and Segmentation for Generalized Surgical Instrument Segmentation on Unlabelled Data} %\thanks{Supported by organization x.}}
\author{Megha Kalia \inst{1} \thanks{M. Kalia and T. Aleef contributed equally to the manuscript} \and
Tajwar Abrar Aleef\inst{1} \samethanks  \and
Nassir Navab\inst{2}\and 
Septimiu E. Salcudean\inst{1}
}

% \footnotemark[1]
%
\authorrunning{M. Kalia  \& T. Aleef  et al.}
\institute{Electrical and Computer Engineering,
        University of British Columbia, 2332 Main Mall, Vancouver, BC Canada\\
\email{\{mkalia, tajwaraleef, tims\}@ece.ubc.ca}\\
\and
Computer Aided Medical Procedures, Technical University of Munich, Boltzmannstr. 3, 85748 Garching bei München, Germany \\
\email{nassir.navab@tum.de}}

\maketitle              % typeset the header of the contribution
\begin{abstract}
Surgical instrument segmentation for robot-assisted surgery is needed for accurate instrument tracking and augmented reality overlays. Therefore, the topic has been the subject of a number of recent papers in the CAI community. Deep learning-based methods have shown state-of-the-art performance for surgical instrument segmentation, but their results depend on labelled data. However, labelled surgical data is of limited availability and is a bottleneck in surgical translation of these methods. In this paper, we demonstrate the limited generalizability of these methods on different datasets, including human robot-assisted surgeries. We then propose a novel joint generation and segmentation strategy to learn a segmentation model with better generalization capability to domains that have no labelled data. The method leverages the availability of labelled data in a different domain. The generator does the domain translation from the labelled domain to the unlabelled domain and simultaneously, the segmentation model learns using the generated data while regularizing the generative model. 
We compared our method with state-of-the-art methods and showed its generalizability on publicly available datasets and on our own recorded video frames from robot-assisted prostatectomies. Our method shows consistently high mean Dice scores on both labelled and unlabelled domains when data is available only for one of the domains.

\end{abstract}
%
%
%

% \footnote[1]{M. Kalia and T. Aleef contributed equally to the manuscript }

\section{Introduction}

%Robot-Assisted Surgery (RAS) has been rapidly adopted as a standard treatment for many medical conditions. Surgery through an endoscope and robotic interface, paves way for wide computer vision and Augmented Reality (AR) applications to understand a surgical scene and enhance a surgeon's vision.
Surgical instrument segmentation is fundamental to Augmented Reality (AR) in image-guided robot-assisted surgery (RAS) \cite{pauly2014relevance} and has been an active topic of research, with convolutional neural network (CNN)-based methods surpassing prior methods by a significant margin \cite{attia2017surgical,pakhomov2019deep,iglovikov2018ternausnet}. CNN-based methods depend on the availability of annotated surgical data, which may be difficult to obtain \cite{maier2017surgical}. Their performance has been reported for publicly available {\em ex vivo} and porcine {\em in vivo} RAS surgeries, but not in human RAS.

\begin{figure}[t]
\centering
\includegraphics[width=\textwidth]{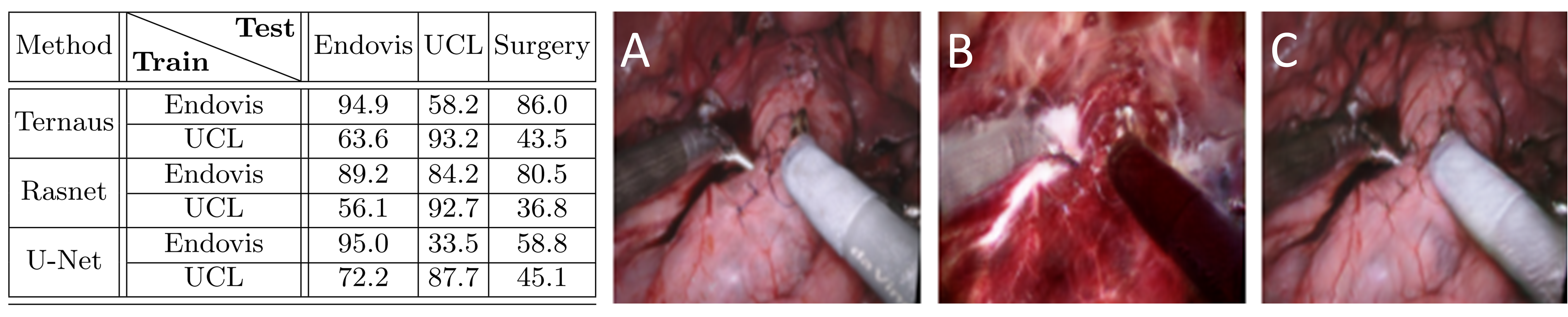}
\caption{(Left) Table showing the limited generalizability of state of the art (SOTA) methods across domains. Mean Dice scores are shown for different methods on datasets \textit{Endovis}, \textit{UCL}, and \textit{Surgery}. (Right) Figures showing the problem with cycleGAN where the intermediate generated output can be unrealistic while overall cycle consistency loss is low. Figures A, B, and C show the original, translated, and reconstructed domain.} \label{fig1}
\end{figure}

%Contrary to real human surgeries, these datasets have mostly clean surgical instruments and in visually differ significantly from actual human RAS surgeries. 
%
Recently, many generative approaches have been proposed to mitigate the problem of limited clinical labelled data \cite{ross2018exploiting,colleoni2021robotic,lin2020lc}. 
For laparoscopic instrument segmentation, \cite{ross2018exploiting} proposed a generative adversarial network (GAN) based method to use a small amount of labelled data.
%The pre-trained network was fine tuned on a small labelled data in the source domain for instrument segmentation. 
%Although the method requires much less labelled data, it still is dependent on labelled data which is resource intensive. 
In \cite{lin2020lc}, labelled data from cadaver surgery was transferred to {\em in vivo} surgery. 
Then a separate segmentation model was trained using either the translated cadaver data or translated {\em in vivo} data to the cadaver domain. 
In \cite{colleoni2021robotic}, an image-to-image (I2I) mapping of simulated to real surgical instruments was proposed, with blending into the camera background. 
In the above methods, the translated data was used to train a segmentation model.  
Finding validated quantitative metrics for the quality of translated data is difficult and is the topic of on-going research \cite{kynkaanniemi2019improved,zhou2019hype}. In many cases, the generative models can change the surgical instruments' shapes and introduce artefacts while the overall accuracy goes down [Figure~\ref{fig1}]; both of which are undesirable for clinical application. Hence, a segmentation strategy leveraging the power of generative models to alleviate the problem of unlabelled clinical data while addressing the predominant current challenges of generative models is imperative.

Therefore, in the current paper, we present a joint unpaired I2I mapping and segmentation strategy for better generalizability of a surgical instrument segmentation model to a domain with no labelled data. 
The generative and segmentation models are trained together and reach convergence in a synergistic manner. The generative model maps from a source domain with labelled data to a target domain with unlabelled data with constant feedback from the segmentation model. The segmentation model trains in parallel on the generated target images and on the labelled source images. The convergence criterion of this joint-system is the segmentation quality. The segmentation model also regularizes the generative model that can otherwise change the shape of the surgical instruments during the I2I mapping. 
%To provide an intermediate supervision and shape constraint during the generation process, we enforce an additional latent space-based shape loss on the target and source domain generators. %REPEATED BELOW
We call our method coSegGAN. The closest method to it is presented in \cite{lin2020lc}. However, unlike in \cite{lin2020lc}, our segmentation model is not pre-trained. It provides feedback to the generators as it learns using the generated data, thus seeing much more varied data. Unlike prior work, we provide an explicit shape constraint on latent space to provide intermediate supervision during generative training. Through evaluation on real surgical sequences and publicly available datasets, we show that coSegGAN has better generalizability than existing methods. 
%While we report results using a specific segmentation model, the strategy can easily be adopted for any segmentation model thus making the technique applicable to various surgical scenarios. 
To the best of our knowledge, this is the first method that segments surgical instruments with no labelled data by jointly training the generative and segmentation model as a joint feedback system to perform an I2I mapping between the labelled and the unlabelled domain. 

% \begin{table}
% \begin{tabular}{|c||c||c|c|c| }
% \hline
%   Method & \diagbox{\textbf{Train}}{\textbf{Test}}& Endovis & UCL & Surgery\\\hline\hline
%  \multirow{2}{*}{Ternaus} & Endovis & 94.9 &58.2 &86.0\\\cline{2-5}
%  & UCL & 63.6 &93.2 & 43.5\\\hline
%  \multirow{2}{*}{RASnet} & Endovis& 89.2 &84.2 &80.5\\\cline{2-5}
%  & UCL &56.1 &92.7 & 36.8\\\hline
%  \multirow{2}{*}{U-Net} & Endovis & 95.0 & 33.5 &58.8\\\cline{2-5}
%  & UCL & 72.2 &87.7 & 45.1\\
%  \hline
%  \hline
% \end{tabular}
% \end{table}

\section{Methods} \label{sec:Methods}
% We propose two methods. First method performs an I2I mapping between the source domain with labelled data and the target domain with unlabelled data and jointly trains a segmentation model with source and generated target domain data. We call this method coSegGAN in the rest of the paper. 
%To distinguish our method from other methods we will call our method coSegGAN in rest of the paper. 

\subsection{Network Details}
The generative part of coSegGAN uses a cycleGAN like architecture with two generators and two discriminators \cite{zhu2017unpaired}. Let $x_{ai}$ and $x_{bi}$ denote the two $i^{th}$ images in the domains $\psi_{A}$ and $\psi_{B}$, respectively, and $x_{a}$ and $x_{b}$ denote the set of all images in domains $a$ and $b$, respectively. $y_{ai}$ denotes corresponding individual label for $i^{th}$ $x_{ai}$ image and $y_{a}$ is the set of all such labels. The $G_{A}$ and $G_{B}$ are the two generators estimating the mappings, $G_{A}: x_{b\rightarrow a}$ and $G_{B}: x_{a\rightarrow b}$, respectively. The discriminator $D_{A}$ is responsible to discriminate between given true images in domain $\psi_{A}$ and generated images $G_{A}(x_{b})$. Similarly, $D_{B}$ is responsible to discriminate between the true domain $\psi_{B}$ and generated images $G_{B}(x_{a})$. Both $G_{A}$ and $G_{B}$ have a U-Net-like architecture \cite{ronneberger2015u} with a contracting and expanding path. The contracting path consists of four $4~ \times~ 4$ convolutional layer with stride $2$ + Leaky ReLu + Instance normalization \cite{ulyanov2016instance} blocks where in each subsequent block the output is halved and the channel numbers are doubled. The expanding path consists of three blocks with each block consisting of an up-sampling layer + $4~ \times~ 4$ convolution with a stride of $1$ + ReLu activation + Instance normalization. The output of each block was concatenated with the low-level features from the contracting path by skip connections and then passed as an input to the next block. The output of the final block was passed though a convolutional layer followed by a tanh activation. For the discriminator, we used a patchGAN similar to \cite{zhu2017unpaired}. For the segmentation model ($S$) in coSegGAN we used the original U-Net architecture but with $16$ base filters to prevent over-fitting and to reduce computation. This did not decrease the performance of segmentation when compared to the original U-Net, as determined empirically. 

\begin{figure}[t]
\includegraphics[width=\textwidth]{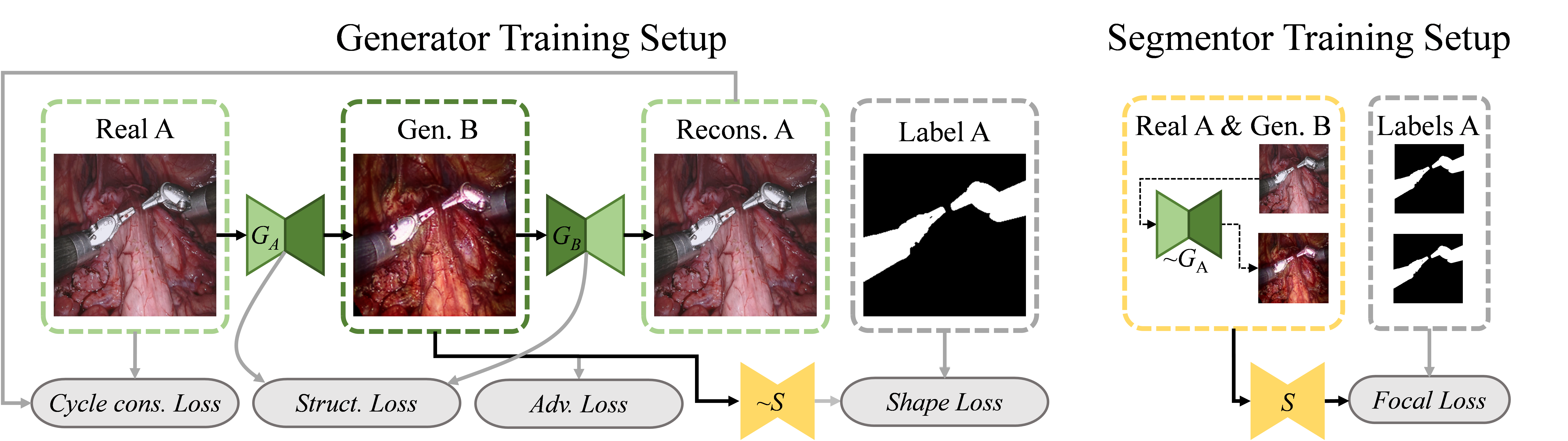}
\caption{Overview of the training setup for the generation and segmentation side. The diagram on the left shows the A to B mapping side of the cycleGANs that is modified to incorporate shape loss and structural loss during training. The grey blocks indicate the different losses used for updating the weights of the generator. On the right side, the input and loss used for the segmentation model training are shown. Any networks indicated by `$\sim$' in this figure indicates it has frozen weights.} \label{fig2}
\end{figure}

\subsection{Training Strategy}
We trained the generators, discriminators \& the segmentation model in an alternative fashion. In the first run, the weights through the generators, $G_{A}$ and $G_{B}$ were back-propagated while freezing the weights of the discriminators and the segmentation model. Then in the next run, the discriminators as well as the segmentation model, $D_{A}$, $D_{B}$, \& $S$ were trained and updated. For training $S$, both $x_{a}$ and $G_{B}(x_{a})$ were fed as the input. Since the generated images are translated versions of the real image, the corresponding labels for $G_{B}(x_{a})$ are the same as $x_{a}$. Note that $S$ is seeing different variations of the generated target domain images in every epoch because the generators and $S$ are learning in parallel. While the quality of the I2I mapping from the generators increases, the quality of the images seen by $S$ also increases. Details can be seen in  Fig.~\ref{fig2}.

\subsection{Loss Functions}

\textbf{Segmentation Model}
In order not to overwhelm the loss with the higher number of background pixels, we used an $\alpha$-balanced variant of focal loss, $\mathcal{L}_{foc}$ \cite{lin2017focal}, a modification of cross-entropy, where the $\gamma$ factor controls the contribution of high-probability samples in the loss calculation. We used the hyper-parameters $\gamma$ and $\alpha$ as $2.0$ and $0.25$, respectively. The total segmentation loss, $\mathcal{L}_{seg}$, is 
\begin{equation}
    \mathcal{L}_{seg} = \mathcal{L}_{foc}\left(x_{a}, y_{a}\right) + \mathcal{L}_{foc}\left(G_{B}\left(x_{a}\right), y_{a}\right) \label{eq:1}
\end{equation}
%
% \begin{equation}
%     \mathcal{L}_{seg} = - \alpha_{t}(1 - p_{t})^{\gamma}\log(p_{t})
% \end{equation}
%
\textbf{Generative Model}
For cycleGAN we used an adversarial loss, $\mathcal{L}_{GAN}$, and a pixel-level cycle consistency loss $\mathcal{L}_{cyc}$ proposed in \cite{zhu2017unpaired}. Although $\mathcal{L}_{cyc}$ reduces the number of possibilities when mapping across domains and regularizes the cycleGAN, it does not suffice to preserve the higher-level semantics in the image. This can change the shapes of surgical instruments during the translation, which is not desirable. Therefore we included feedback from the segmentation model in the total generative loss. This penalizes the generation of unrealistic surgical instrument shapes in $G_{B}\left(x_{a}\right)$. Since we are interested in the mapping from $x_{a}$ to $x_{b}$, which later is fed as an input to the segmentation model, we included this constraint only on the generator $G_{B}$. This shape preservation loss, $\mathcal{L}_{shape}$, is
\begin{equation}
    \mathcal{L}_{shape} = \mathcal{L}_{foc}\left(G_{B}\left(x_{a},\right) y_{a}\right) \label{eq:2}
\end{equation}
%
% \begin{equation}
%     \mathcal{L}_{shape} = \E\left[\norm{S(G_{A}(G_{B}(x_{a}))) - y_{a}}_{1}]\right + \E\left[\norm{S(G_{B}(x_{a})) - y_{a}}_{1}]\right \label{eq:2}
% \end{equation}
%
In cycleGAN models, $\mathcal{L}_{cycTotal}$ is sum of two cycle consistency losses such that, $\mathcal{L}_{cycTotal} = \mathcal{L}_{cyc}(x_{a}, G_{A}(G_{B}(x_{a}))) + \mathcal{L}_{cyc}(x_{b}, G_{B}(G_{A}(x_{b})))$. These losses enforce pixel level constraints between the original inputs $x_{a}$ and $x_{b}$ and reconstructed outputs $ G_{A}(G_{B}(x_{a}))$ and $ G_{A}(G_{B}(x_{a}))$, where the two GANs are optimized together. There is no intermediate supervision after each generative step $G_{A}: x_{b\rightarrow a}$ and $G_{B}: x_{a\rightarrow b}$. Thus $G_{A}$ and $G_{B}$ can produce unrealistic images while the total $\mathcal{L}_{cyc}$ is reduced (Shown in Fig.~\ref{fig1}, (right)).
In particular, the mapping across domains should change only the `appearance' of the scene while retaining the domain-invariant structural elements. 
To preserve the structural properties of the scene across domains, we introduce an explicit, intermediate, feature level, latent space loss. This latent space loss, and the total generated loss, are:
\begin{eqnarray}
    && \mathcal{L}_{structure} = \E\left[\norm{e_{A}(x_{a}) - e_{B}(G_{B}(x_{a})) }_{1}\right] + \E\left[\norm{e_{B}(x_{b}) - e_{A}(G_{A}(x_{b})) }_{1}\right] \label{eq:2} \\
   && \mathcal{L}_{generator} = \lambda_{1}\mathcal{L}_{GANTotal}+\lambda_{2} \mathcal{L}_{cycTotal}+ \lambda_{3} \mathcal{L}_{shape} + \lambda_{4}\mathcal{L}_{structure} +  \lambda_{5}\mathcal{L}_{I}\,\,. \label{eq:3}
\end{eqnarray}
where, $e_{A}$ and $e_{B}$ are encoders in $G_{B}$ and $G_{A}$, respectively, $\mathcal{L}_{GANTotal} = \mathcal{L}_{GAN}(G_{B}, D_{B}, x_{a}, x_{b}) + \mathcal{L}_{GAN}(G_{A}, D_{A}, x_{b}, x_{a})$ and $\mathcal{L}_{I}$ is the identity mapping loss as given in \cite{zhu2017unpaired}. Values of $\lambda_{1}$, $\lambda_{2}$, $\lambda_{3}$, $\lambda_{4}$, and $\lambda_{5}$ are $1$, $10$, $1$, $5$, and $1$, respectively. These values were tuned during the hyper-parameter tuning phase.

% {L}_{cyc}(x_{a},G_{A}(G_{B}(x_{a})))

%TODO add equation of cycle consistency loss with the U-Net loss

\subsubsection{Training Details and Hyper-Parameters}
For training and testing our models, we use Tensorflow \& Keras API on a NVIDIA Tesla V100 GPU (16GB). For training the proposed models, we used a batch size of 8, and Adam optimizer with $\beta_{1}$ and $\beta_{2}$ of $0.9$ and $0.999$, respectively, with a learning rate of $10^{-3}$. We trained our models for $100$ epochs (approximately $12$ hrs) and saved weights of the segmentation model with the highest validation Dice score \cite{zou2004statistical}. Code is available at: \textit{anonymous website}

\section{Experiments}

\subsubsection{Datasets}:
\noindent
\textbf{Endovis Challenge, $2017$, \textit{in-vivo} dataset \cite{allan20192017}}
It is a porcine surgery procedure with a training set consisting of $8$ videos of $225$ frames each and a test set consisting of $8$ videos of $75$ frames and $2$ videos of $300$ frames each. We used $6$ videos for training and $2$ videos for validation from the training set. We used $8$ videos from the test set for testing that was not used for validation. In the paper, we refer to this dataset as \textit{Endovis}.

\noindent
\textbf{UCL \textit{ex-vivo} dataset \cite{colleoni2020synthetic}}
The dataset consists of $14$ videos with different animal tissues as background. Similarly to \cite{colleoni2020synthetic}, we used $8$, $2$ and $4$ videos for training, validation and testing, respectively.

\noindent
\textbf{Prostatectomy dataset}
We prepared the training dataset from $5$ videos of robot-assisted radical prostatectomy procedures with the da Vinci Si surgical system from Anonymous Hospital. We manually selected $1327$ frames to isolate surgical instruments from other visible objects in the surgical field of view. These frames do not have corresponding labels. To evaluate the performance of the various methods on actual surgical data, we prepared a test set of $182$ frames taken from $4$ different surgeries independent from the training set. The test data represents approximately $12\%$ of the entire surgical data used. We manually labelled surgical instruments in these frames only for the purpose of testing coSegGAN and existing methods. All the frames were center cropped to give a final size of $ 721 \times 503 ~pixels$. We will refer to this dataset as \textit{Surgery} in the rest of the paper. Ethics to collect data was obtained from the Institutional Clinical Research Ethics Board. For all three datasets, we resized the frames to $256 \times 256$ to accelerate the computation.

\subsubsection{Evaluation} 
We compared coSegGAN with Ternausnet, the best performing method in the Endovis Challenge \cite{allan20192017} for binary segmentation and RASnet, reporting a mean $94.65\%$ Dice coefficient on Endovis. 
For a fair comparison to coSegGAN, we performed data augmentation with cycleGAN architecture given in section \ref{sec:Methods}. cycleGAN model was run for $50$ epochs in all cases as it converged in $50$ epochs. After cycleGAN I2I translation from source (with labels) to target domain, the SOTA segmentation models were trained with both the translated and original domain data. We also performed an ablation experiment comparing coSegGAN with and without the proposed $\textbf{L}_{structure}$ loss. We refer to RASnet, Ternausnet, and our U-Net variant with focal loss, trained using the augmented data generated from a separate cycleGAN (unlike our joint strategy) as $RASnet+$, $Ternausnet+$ and $U\text{-}Net_{FL}+$ respectively. The coSegGAN network without $\textbf{L}_{structure}$ is called $coSegGAN-$. We performed evaluation of four combinations of datasets for labelled and unlabelled domains. For ease of reporting, we refer to \textit{Endovis} (labelled) + \textit{Surgery} (Unlabelled),  \textit{UCL} (labelled) + \textit{Surgery} (Unlabelled), \textit{Endovis} (labelled) + \textit{UCL} (Unlabelled), and \textit{UCL} (labelled) + \textit{Endovis} (Unlabelled) data combinations as case $1$, case $2$, case $3$, and case $4$, respectively. Since, we want to quantify the generalizability of our method across labelled and unlabelled domains, for a particular dataset combination, we also calculated an absolute difference in the Dice scores, $\Delta ~Dice$, between labelled domain $A$ and unlabelled domain $B$. The lower the $\Delta ~Dice$, the higher is the generalizability between domains (see Table~\ref{tab1}).

\begin{table}[t]%[!htbp]
\centering
\caption{Comparison of Mean Dice Scores of coSegGAN with Existing Methods}\label{tab1}
\begin{tabular}{|p{1.8cm}||p{1.3cm}||p{1.5cm}|p{1.5cm}||p{1.5cm}|p{1.5cm}||p{1.5cm}| }
% \multicolumn{1}{|c||}{\textbf{Method}} &
% \multicolumn{1}{c||}{\textbf{Training Dataset}} & 
% \multicolumn{3}{c|}{\textbf{Test Dataset}} \\
\hline
  Method & &DomainA & DomainB (Unlabelled) & 
  Domain A & DomainB (Unlabelled) &  $\Delta~Dice$ \\\hline\hline
       
\multirow{4}{*}{$coSegGAN$} & case 1& Endovis & Surgery &  $\textbf{93.7\%}$ & $\textbf{92.8\%}$ & $\textbf{0.9\%}$ \\\cline{2-7}
                 & case 2&UCL & Surgery & $91.1\%$ & $74.3\%$ & $16.8\%$ \\\cline{2-7}
                 & case 3 & Endovis & UCL & $\textbf{93.2\%}$ & $\textbf{90.0\%}$ & $\textbf{3.2\%}$ \\\cline{2-7}
                 & case 4&UCL & Endovis & $93.5\%$ & $79.4\%$ & $14.1\%$\\\hline \cline{1-7}

\multirow{4}{*}{$RASnet+$} & case 1&Endovis & Surgery & $88.3\%$ & $78.1\% $& $10.2\%$\\\cline{2-7}
                             &case 2& UCL & Surgery &  $92.3\%$ & $47.8\%$ & $44.5\%$ \\\cline{2-7}
                             &case 3& Endovis & UCL & $88.4\%$ & $83.3\%$ & $5.1\%$\\\cline{2-7}
                             & case 4&UCL & Endovis & $92.4\%$ & $66.8\%$ & $25.6\%$\\\hline\cline{1-7}

\multirow{4}{*}{$Ternaus+$} & case 1 & Endovis & Surgery & $94.2\%$ & $88.7\%$ & $5.5\%$\\\cline{2-7}
                             & case 2&UCL & Surgery & $95.8\%$  & $46.0\% $ & $49.8\%$ \\\cline{2-7}
                             & case 3&Endovis & UCL & $93.3\%$ & $41.7\%$ & $51.6\%$ \\\cline{2-7}
                             & case 4&UCL & Endovis & $93.4\%$ & $55.0\%$ & $38.4\%$ \\\hline\cline{1-7}

\multirow{4}{*}{$U-Net_{FL+}$} & case 1 & Endovis & Surgery & $91.8\%$ & $58.0\%$ & $33.8\%$ \\\cline{2-7}
                             & case 2&UCL & Surgery & $93.2\%$ & $36.0\% $ & $57.2\%$ \\\cline{2-7}
                             & case 3&Endovis & UCL & $83.9\%$ & $23.3\%$ & $60.7\%$\\\cline{2-7}
                             & case 4&UCL & Endovis & $74.6\%$ & $56.5\%$ & $18.1\%$ \\\hline \cline{1-7}

% \multirow{4}{*}{U-Net_{FL}+ with HISTEQ AND AUG IN TRAINING} & case 1&Endovis & Surgery &93.4 &84.1 &77.0\\\cline{2-7}
%                              & case 2&UCL & Surgery &93.7 &45.6&48.1 \\\cline{2-7}
%                              & case 3&Endovis & UCL & 94.7& 81.8&12.9\\\cline{2-7}
%                              & case 4&UCL & Endovis &92.6 &77.0&15.6 \\\hline\hline

 \multirow{4}{*}{$coSegGAN--$} &case 1& Endovis & Surgery &  $94.1\%$ & $92.3\%$ & $\textbf{1.8\%}$ \\\cline{2-7}
                             &case 2& UCL & Surgery & $93.5\%$ & $74.5\%$ & $19.0\%$  \\\cline{2-7}
                             &case 3& Endovis & UCL & $93.3\%$ & $90.8\%$ & $\textbf{2.5\%}$ \\\cline{2-7}
                             & case 4&UCL & Endovis & $94.2\%$ & $74.4\%$ & $19.8\%$\\\hline\cline{1-6}

 \hline
 
 \hline
\end{tabular}
\end{table}

% \multirow{2}{*}{RASnet} & Endovis & - &89.2 &84.2 &80.5\\\cline{2-6}
%                              & UCL & - &56.1 &92.7 & 36.8\\\hline \cline{1-6}

% \multirow{2}{*}{Ternaus} & Endovis & - & 94.9 &58.2 &86.0\\\cline{2-6}
%                              & UCL & - &63.6 &93.2 & 43.5\\\hline\cline{1-6}

% \multirow{2}{*}{U-Net} & Endovis & - &95.0 & 33.5 &58.8\\\cline{2-6}
%                              & UCL & - &72.2 &87.7 & 45.1\\

\section{Results and Discussion}

For case $1$, the proposed coSegGAN network gave significantly higher Dice score ($92.8\%$) on unlabelled domain B (\textit{Surgery}) when compared to $RASnet+$, $Ternausnet+$ and $U\text{-}Net_{FL}+$ which have Dice scores of $78.1\%$, $88.7\%$, and $84.1\%$, respectively. 
For case $2$ as well, the Dice score for coSegGAN on unlabelled domain (\textit{Surgery}) is $74.3\%$ while $RASnet+$, $Ternausnet+$, and $U\text{-}Net_{FL}+$ have lower Dice scores of $47.8\%$, $46.0\%$, and $45.6\%$, respectively. Similarly, for case $3$, the Dice score for coSegGAN on unlabelled data (\textit{UCL}) is $90\%$, which is higher than $RASnet+$, $Ternausnet+$ and $U\text{-}Net_{FL}+$ with Dice scores of $83.3\%$, $41.7\%$, and $81.8\%$, respectively. For case $4$, the Dice score for coSegGAN on unlabelled \textit{Endovis} data is $79.4\%$ which, similar to other cases, is higher than the rest of the methods; Dice scores of $RASnet+$, $Ternausnet+$ and $U\text{-}Net_{FL}+$ being $66.8\%$, $55.0\%$ and $56.5\%$, respectively. 
% coSegGAN's significantly higher Dice score on unlabelled data when compared to methods with two step augmentation and segmentation, shows its ability to generate correct representations that can further be learned by segmentation model.

The $\Delta~ Dice$, for coSegGAN for case $1$ is much lower $0.9\%$ while for $RASnet+$, $Ternausnet+$, and $U\text{-}Net_{FL}+$ it is $10.2\%$, $5.5\%$, and  $33.8\%$, respectively. For case $2$, $\Delta~ Dice$ for coSegGAN is $16.8\%$, while for $RASnet+$, $Ternausnet+$ and $U\text{-}Net_{FL}+$ it is $44.5\%$, $49.8\%$ and $57.2\%$, respectively. For case $3$, $\Delta~ Dice$, for coSegGAN is $3.2\%$, which is much lower than $RASnet+$, $Ternausnet+$ and $U\text{-}Net_{FL}+$ with $\Delta~ Dice$ of $5.1\%$, $51.6\%$, and $60.7\%$, respectively. For case $4$, similarly, the $\Delta~ Dice$ for coSegGAN is $14.1\%$ when compared to $RASnet+$, $Ternausnet+$, and $U\text{-}Net_{FL}+$ with $\Delta~ Dice$ of $25.6\%$, $38.4\%$, and $18.1\%$, respectively. Consistently significantly lower $\Delta~ Dice$ on coSegGAN shows its generalizability when compared to all other methods for all the cases.

For coSegGAN, in the cases $1$ and $3$, when the mapping is from \textit{UCL} (labelled) to either \textit{Surgery} or \textit{Endovis}, the $\Delta~ Dice$ is lower than cases $1$ and $2$. This could be because the \textit{UCL} data is an ex-vivo dataset, and does not represent a real surgery, with remarkably different lighting and background. Also, there is only one type of surgical instrument visible in the \textit{UCL} dataset, which might have hindered the mapping to multiple types of instruments.

In the ablation experiment, coSegGAN--, i.e., coSegGAN without the $L_{structure}$, showed comparable performance with coSegGAN, except case $4$, where the performance of coSegGAN is significantly higher (approximately $5\%$) on the unlabelled \textit{Endovis} dataset. coSegGAN-- has higher $\Delta~ Dice$ for all cases except case $4$, showing that with the $L_{structure}$ loss coSegGAN generalizes better to both labelled and unlabelled datasets.

\begin{figure}[t]
\includegraphics[width=\textwidth,height=7.5cm]{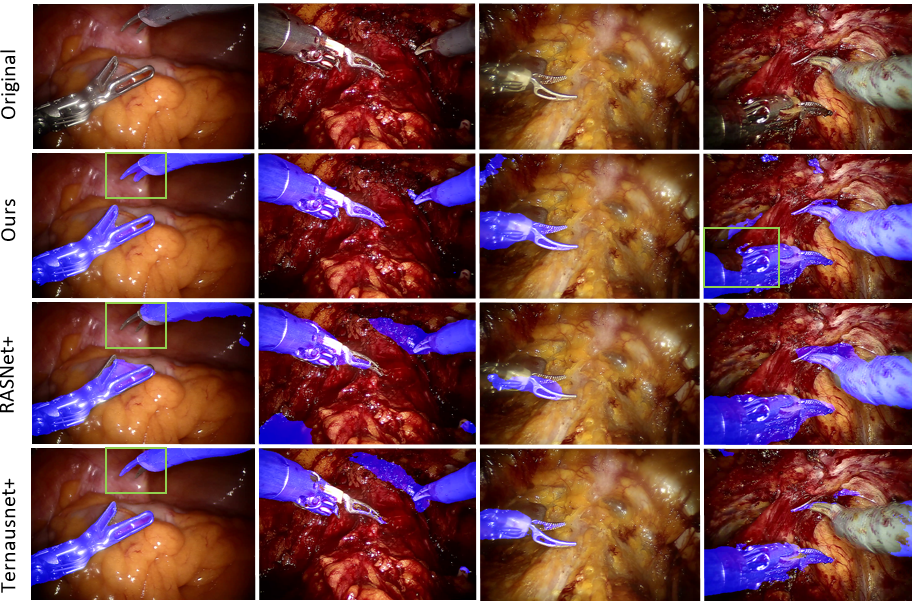}
\caption{Figure showing a qualitative comparison of our method with other methods. It can be seen that overall, our method preserves the shape of the instruments better with fewer false positives. (column $1$) Inset showing preservation of instrument shape in our method. (column $4$) Inset showing a failure case of our method.} \label{fig:robust}
\end{figure}

A qualitative comparison of coSegGAN with other methods for different surgeries can be seen in Fig.~\ref{fig:robust}. As can be seen (column 1), coSegGAN performs better in preserving overall tool structure, with finer details, when compared to other methods. In comparison to $Ternausnet+$ and $RASnet+$, the method also produces fewer false positives [Fig.~ \ref{fig:robust} (column 2)].  Fig.~\ref{fig:robust} (column $3$) shows a failure case of coSegGAN. Although coSegGAN performs better than SOTA methods in identifying tools, it occasionally fails to identify the tool in the presence of blood (Fig.~\ref{fig:robust} (column $2$)), where the region is relatively dark compared to the well-lit image center. 

\section{Conclusion}

We presented a joint generative and segmentation strategy, coSegGAN, that outperforms SOTA methods in its generalization capability to unlabelled domain data. The evaluated SOTA methods use separate I2I mapped data augmentation and segmentation steps. The proposed losses helped to preserve finer tool structure. The method is easy to adapt to other deep learning segmentation methods and thus can significantly improve the existing methods. The method aims to utilize unlabelled surgical data, which is much easier to acquire than labelled data, to improve any instrument segmentation model in a simple yet effective manner. Therefore, coSegGAN has the potential to significantly facilitate surgical translation of current and future surgical tool segmentation methods because it effectively alleviates the problem of unlabelled data. Current testing of coSegGAN has been limited to footage from prostatectomy procedures. A thorough performance analysis for different types of RAS surgeries is part of future work.

\newpage
\bibliographystyle{splncs04}
\bibliography{bibliography}

\end{document}